\documentclass{article}

     \PassOptionsToPackage{numbers, compress}{natbib}

     \usepackage[final]{neurips_2019}

\usepackage[utf8]{inputenc} 
\usepackage[T1]{fontenc}    
\usepackage{hyperref}       
\usepackage{url}            
\usepackage{booktabs}       
\usepackage{amsfonts}       
\usepackage{nicefrac}       
\usepackage{microtype}      

\usepackage{graphicx}
\graphicspath{{figures/}}
\usepackage{subcaption}
\usepackage{multirow}

\title{Generating an Explainable ECG Beat Space With Variational Auto-Encoders}
\author{%
	Tom Van Steenkiste, Dirk Deschrijver, Tom Dhaene\\
	Ghent university - imec, IDLab\\
	Technologiepark-zwijnaarde 126\\
	9052 Gent, Belgium \\
	\texttt{tomd.vansteenkiste@ugent.be} \\
}

\begin{document}

\maketitle

\begin{abstract}
	Electrocardiogram signals are omnipresent in medicine. A vital aspect in the analysis of this data is the identification and classification of heart beat types which is often done through automated algorithms. Advancements in neural networks and deep learning have led to a high classification accuracy. However, the final adoption of these models into clinical practice is limited due to the black-box nature of the methods. In this work, we explore the use of variational auto-encoders based on linear dense networks to learn human interpretable beat embeddings in time-series data. We demonstrate that using this method, an interpretable and explainable ECG beat space can be generated, set up by characteristic base beats.
\end{abstract}

\section{Introduction}
Electrocardiogram (ECG) measurements provide essential information for a wide range of medical applications. An important aspect of analyzing the ECG data is detecting heart beats and classifying each beat per type. The amount of beats to be analyzed can quickly become large and human-based classification is a time-consuming task. To aid in this process, automated approaches have been investigated. Many machine learning methods with high accuracies have been proposed. Recent works present neural networks and deep learning techniques~\cite{al2016deep,acharya2017automated,yildirim2018arrhythmia}. However, the adoption of these models into clinical practice is limited due to the lack of model interpretability which is crucial to ensure trustworthiness of the results~\cite{vellido2012making}.

A simple method to create an explainable machine learning model is constructing a standard rule-based classifier. However, with this approach, the powerful predictive capabilities of neural networks and deep learning cannot be exploited. To improve interpretation of these models, dimensionality reduction techniques have been proposed. Examples include auto-encoder (AE) models which reduce complexity by forcing the model to use a lower-dimensional embedding of the input data. Still, complex interactions across the individual dimensions of the learned embedding exist.

In image processing, complex entangled embeddings can be disentangled using disentangled variational auto encoders ($\beta$-VAE)~\cite{higgins2016beta}. These models are capable of learning disentangled generative embeddings by forcing the model to represent the information in as few dimensions as possible, while using a probabilistic interpretation of the embedding. During training, a generative model is created that allows analysts to measure and see the impact of the position within a specific dimension of the embedding. In doing so, the reason for a specific model decision can be traced back to an embedding that has independent and explainable parameters.

In this work, the use of such a $\beta$-VAE based on a linear dense network is investigated for creating an interpretable and explainable ECG beat embedding based on time-series data. The method is used to create a subspace of normal and paced beats from the MIT BIH arrhythmia dataset~\cite{moody2001impact} set up by a characteristic set of base beats. This extended abstract is based on previously published work~\cite{vansteenkiste2019interpretable}.

\section{Variational Auto-Encoders}
\label{sec:vae}
An AE is an unsupervised deep learning model used for creating a lower dimensional embedding, also known as latent representation, of the input data. This embedding is subsequently used in, among others, classification, detection or compression algorithms. In recent studies, AE models were used for classification~\cite{xia2018automatic,ochiai2018arrhythmia} and compression of ECG data~\cite{yildirim2018efficient}. 

\begin{figure}[!htpb]
	\centering
	\begin{subfigure}[h]{0.3\textwidth}
		\includegraphics[width=\textwidth]{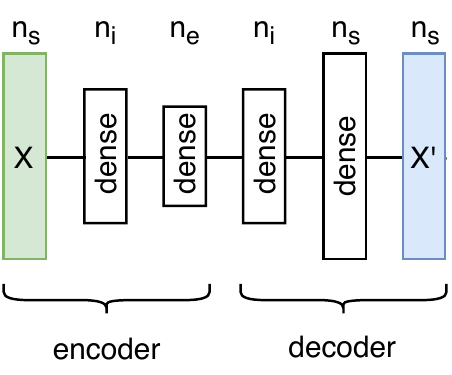}
		\caption{Auto-encoder with input dimensionality $n_s$, intermediate layer of size $n_i$ and embedding of size $n_e$. The input is represented by $X$ and the reconstructed output is represented by $X'$.}
		\label{fig:AEmodel}
	\end{subfigure}
	~
	\begin{subfigure}[h]{0.4\textwidth}
		\includegraphics[width=\textwidth]{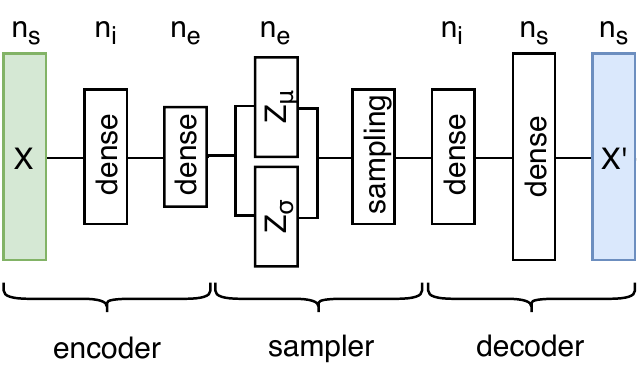}
		\caption{Variational auto-encoder with similar parameters as Fig.~\ref{fig:AEmodel}. The sampler draws a sample from  $\mathcal{N}(Z_{\mu},Z_{\sigma})$.}
		\label{fig:VAEmodel}
	\end{subfigure}
	\caption{Auto-encoder and variational auto-encoder models.}
	\label{fig:}
\end{figure}

A typical AE model consists of two parts: the encoder and decoder, as shown in Fig.~\ref{fig:AEmodel}. The model is then trained using standard deep learning algorithms and a loss function representing the reconstruction loss $L_R$. However, determining the size of the embedding is not straightforward and complex interactions across different dimensions can be created during training.

To get an interpretable embedding, the variational AE (VAE) model can be used~\cite{kingma2013auto}. It transforms regular AE models into probabilistic methods. The embedding layer of Fig.~\ref{fig:AEmodel} is exchanged for two vectors of equal size $Z_{\mu}$ and $Z_{\sigma}$ followed by a sampler drawing a random sample from the distribution $\mathcal{N}(Z_{\mu},Z_{\sigma})$, as shown in Fig.~\ref{fig:VAEmodel}. This random sample is then used by the decoder part.

During training, independence and interpretability of the embedding dimensions is encouraged by the addition of the KL-divergence $D_{KL}$ to the loss function of the model. It is computed between $\mathcal{N}(Z_{\mu_i},Z_{\sigma_i})$ and the standard normal $\mathcal{N}(0,1)$ for each dimension $i$ of the embedding. This encourages the embedding to consist of independent and standard normally distributed dimensions. The effect of this is enhanced in $\beta$-VAE by the addition of a hyperparameter $\beta$ resulting in ${loss} = L_R + \beta D_{KL}$. This hyperparameter balances the latent embedding capacity, also known as channel capacity, with the independence and standard normal distribution constraints~\cite{higgins2016beta}. The resulting model is capable of automatic discovery of independent, interpretable embeddings. More details are presented in ~\cite{higgins2016beta,kingma2013auto}.

\section{Experimental Setup}
\label{sec:experimentalsetup}
To analyze the use of AE and $\beta$-VAE methods to create an interpretable subspace of ECG beats, the MIT-BIH Arrhythmia dataset~\cite{moody2001impact} is used. All patients with paced beats are included and an equal amount of patients with normal beats are added. To accurately test the capabilities of the models, the data is split in a separate training and test set. The patient identifiers for each set are given in Table~\ref{table:patdist}. The annotations included in the database are used to detect and categorize the beats. Only normal and paced beats are included in the experiment.

\begin{table}[!htpb]
	\centering
	\caption{Distribution of patients across train and test set.}
	\label{table:patdist}
	\begin{tabular}{l|ll}
		\hline
		& \textbf{normal} & \textbf{paced} \\ \hline
		\textit{train} & 101, 106        & 102, 104       \\
		\textit{test}  & 103, 105        & 107, 217       \\ \hline
	\end{tabular}
\end{table}

\begin{figure}[!htpb]
	\centering
	\begin{subfigure}[h]{0.23\textwidth}
		\includegraphics[width=\textwidth]{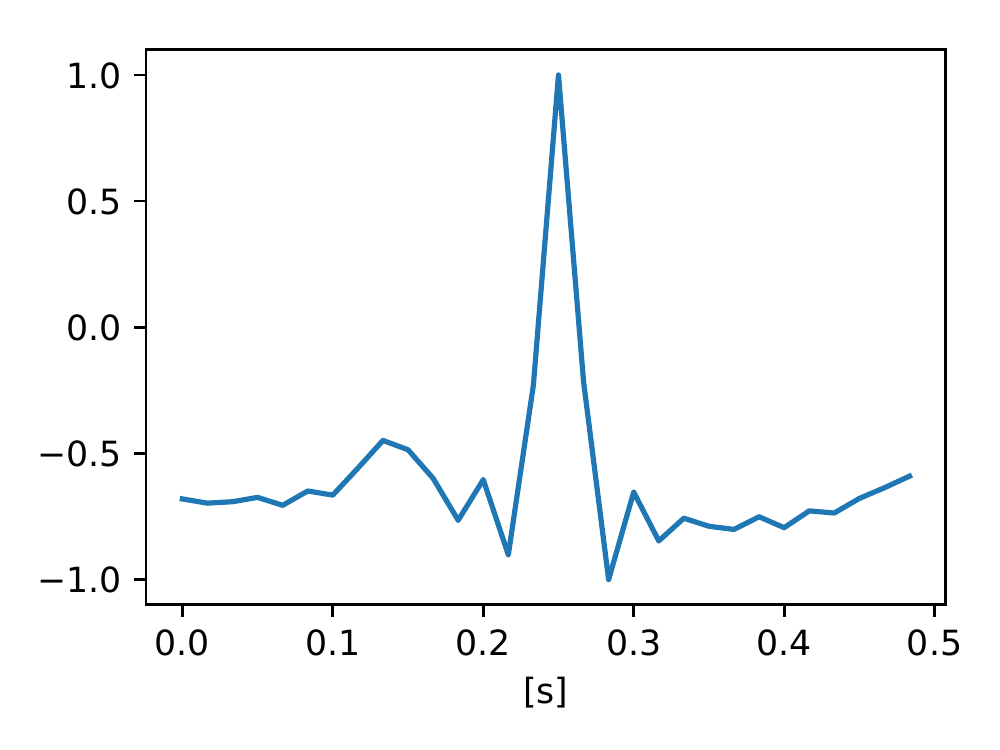}
		\caption{Normal beat - train.}
		\label{fig:dataNtrain}
	\end{subfigure}
	~
	\begin{subfigure}[h]{0.23\textwidth}
		\includegraphics[width=\textwidth]{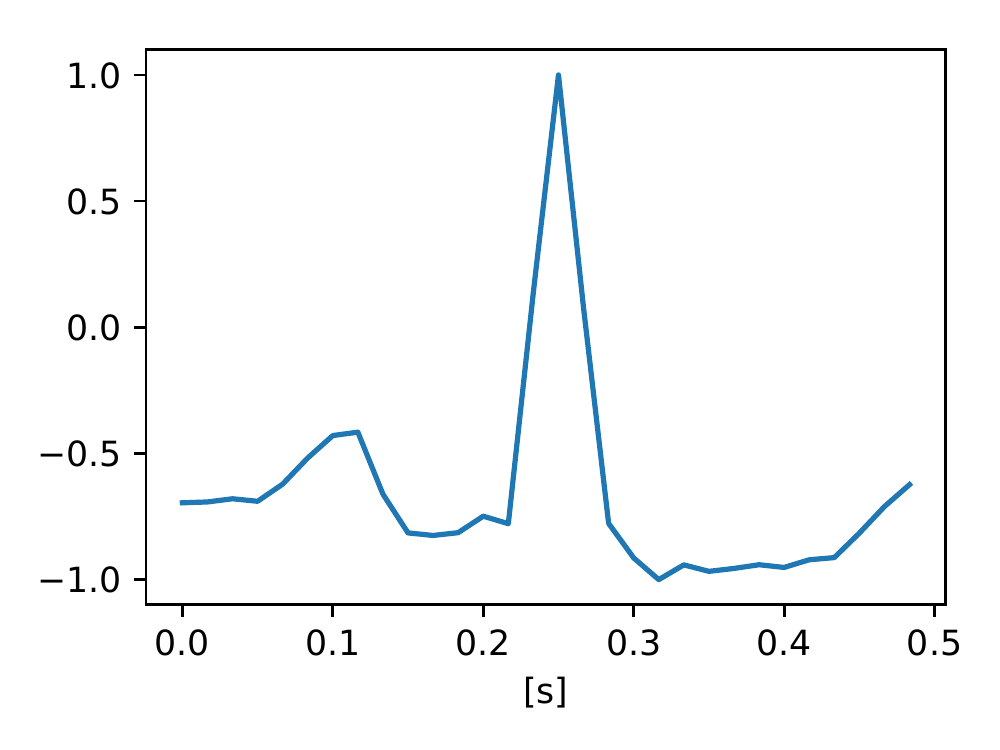}
		\caption{Normal beat - test.}
		\label{fig:dataNtest}
	\end{subfigure}
	~
	\begin{subfigure}[h]{0.23\textwidth}
		\includegraphics[width=\textwidth]{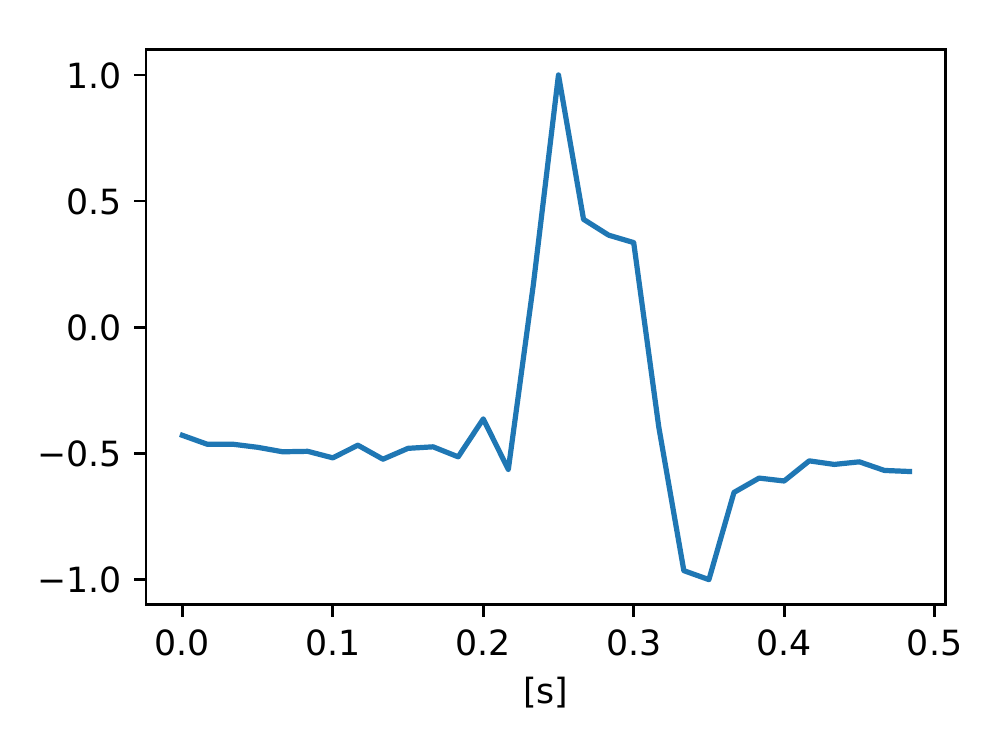}
		\caption{Paced beat - train.}
		\label{fig:dataPtrain}
	\end{subfigure}
	~
	\begin{subfigure}[h]{0.23\textwidth}
		\includegraphics[width=\textwidth]{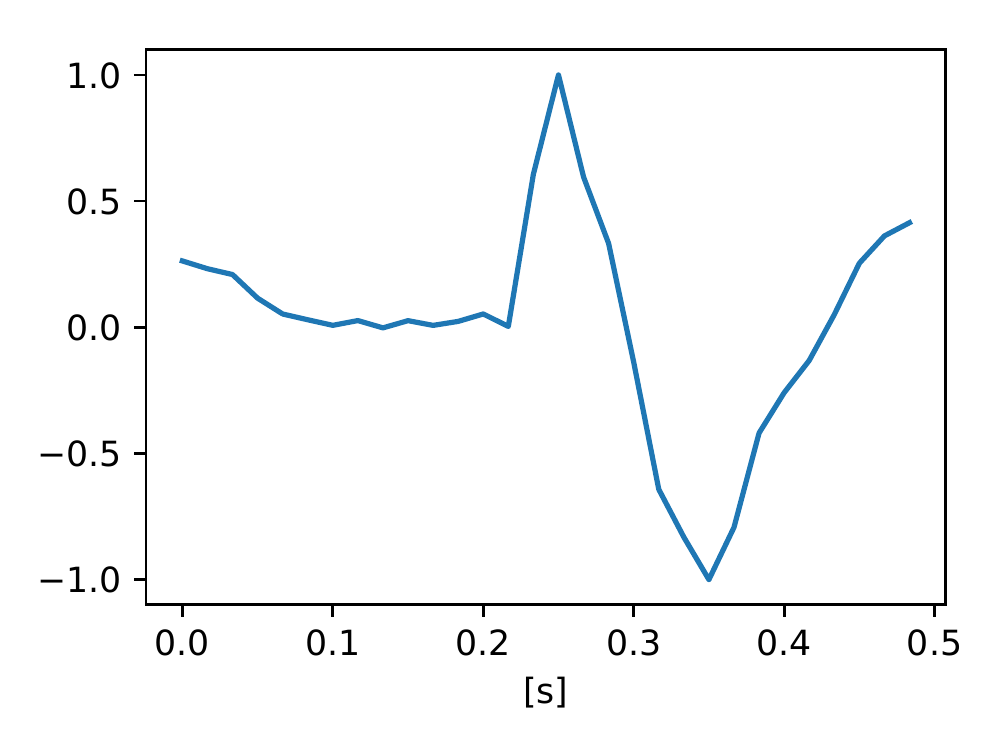}
		\caption{Paced beat - test.}
		\label{fig:dataPtest}
	\end{subfigure}
	\caption{Random samples form train and test dataset used in this analysis. There is a visual and interpretable difference between the beats.}
	\label{fig:data}
\end{figure}

The ECG signal is passed through a fifth-order Butterworth bandpass filter with lower cutoff frequency of 1Hz and upper cutoff frequency of 60Hz for mild noise removal. The epochs of data have a duration of 1 second, sampled at 60Hz, with the beat centered in the epoch. Then, the signal is normalized between $[-1,1]$ and the center 0.5 seconds of data is extracted. This results in 30 samples per epoch (=$n_s$). Examples of the resulting epochs are shown in Fig.~\ref{fig:data}.

Both the $\beta$-VAE as the AE model are constructed with an embedding size $n_e$ of 10 nodes. The intermediate layer has size $n_i$ of 20 nodes. All dense layers have linear activation functions resulting in a linear dense network. Each model is trained for 50 epochs on batches containing 128 samples and is optimized using the AdaDelta~\cite{zeiler2012adadelta} optimizer. The root-mean-squared-error is used to represent the reconstruction loss $L_R$.

\section{Results and Discussion}
\label{sec:resultsanddiscussion}
\begin{figure*}[!htpb]
	\centering
	\begin{subfigure}[h]{\textwidth}
		\centering
		\includegraphics[width=\textwidth]{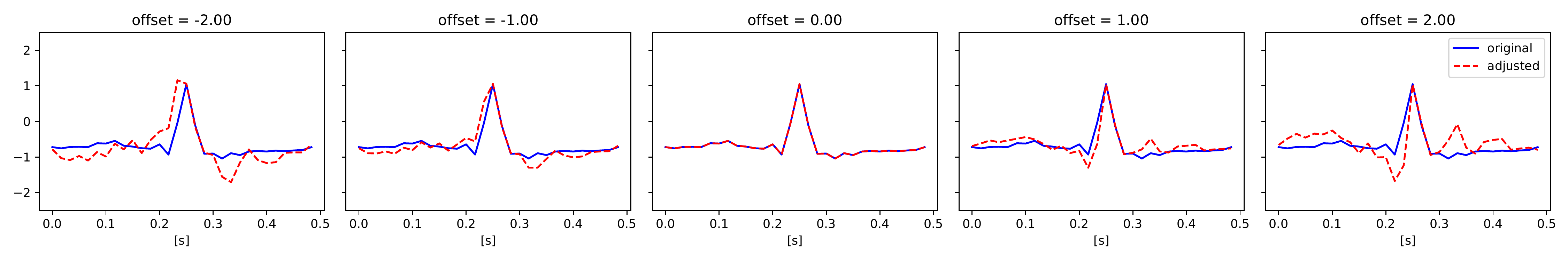}
		\caption{Normal beat perturbed in dimension 4.}
		\label{fig:aepn1}
	\end{subfigure}
	
	\begin{subfigure}[h]{\textwidth}
		\centering
		\includegraphics[width=\textwidth]{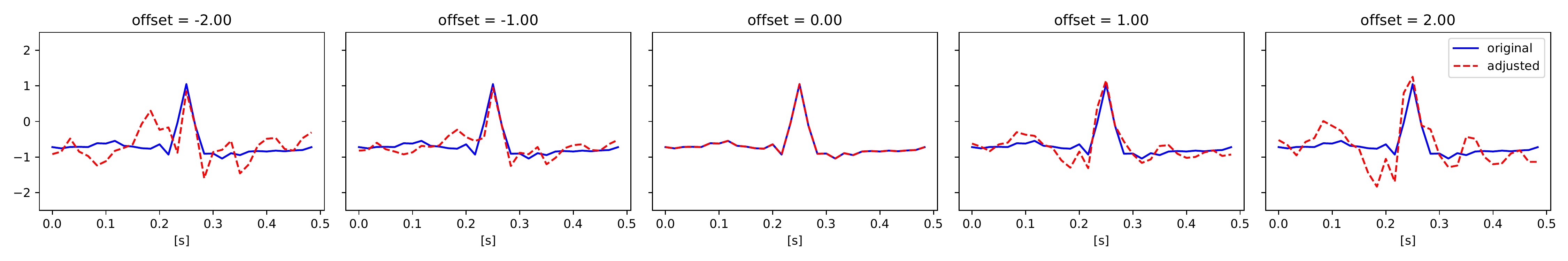}
		\caption{Normal beat perturbed in dimension 7.}
		\label{fig:aepn2}
	\end{subfigure}
	\caption{Perturbing the embedding of an AE model. Perturbing only one dimension generates unrecognizable patterns. Hence, the position within the embedding is not interpretable by human experts.}
	\label{fig:aepertu}
\end{figure*}

Both models are able to create an embedding for the normal and paced beats. The difference between the models is in the explainability of the position within the embedding. This can be analyzed by perturbing each dimension individually as shown in Fig.~\ref{fig:aepertu} for two random dimensions of the AE model. The resulting decoded epochs consist of many peaks and valleys and no longer contain a recognizable beat pattern. No interpretation can be linked with any dimension as the embedding is a complex combination of all 10 dimensions and only makes sense at very specific locations. Changes in the embedding do not always lead to valid samples from the original input distribution $\mathcal{X}$.

\begin{figure*}[!htpb]
	\centering
	\begin{subfigure}[h]{\textwidth}
		\includegraphics[width=\textwidth]{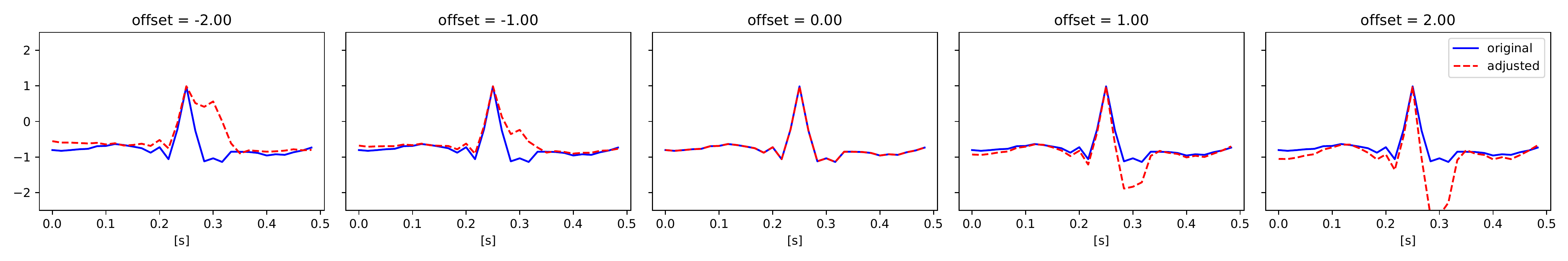}
		\caption{Normal beat perturbed in dimension 0.}
		\label{fig:}
	\end{subfigure}
	
	\centering
	\begin{subfigure}[h]{\textwidth}
		\includegraphics[width=\textwidth]{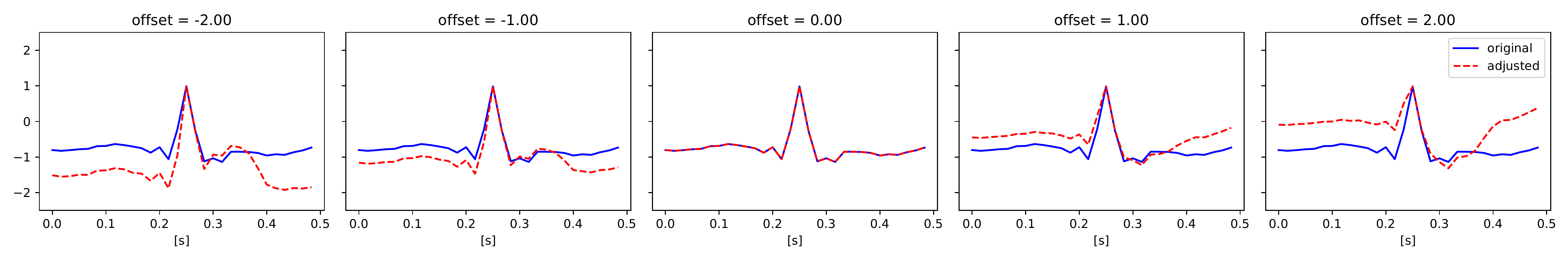}
		\caption{Normal beat perturbed in dimension 9.}
		\label{fig:}
	\end{subfigure}
	\caption{Perturbing the embedding of normal and paced beats using a $\beta$-VAE. An interpretable evolution of the beat can be seen. Human experts can interpret and analyze the physical meaning of the position withing the embedding.}
	\label{fig:vaepert}
\end{figure*}

The $\beta$-VAE model only learned two significant dimensions. These can be identified by inspecting the standard deviation of the embedding dimensions. Now, perturbing one of the dimensions leads to a smooth transition with identifiable beat patterns as shown in Fig~\ref{fig:vaepert}. When comparing the resulting decoded version with the random samples of the database in Fig.~\ref{fig:data}, it is clear that distinct beat shapes are being learned as base for the embedding. The two dimensions now encode a physical shape of the beat and can be changed independently. Any beat can be represented as a combination of these base beats in a learned beat space. The other (insignificant) dimensions do not encode any information and do not lead to changes in the decoded beat. The amount of significant dimensions is influenced by the $\beta$ parameter of the model which balances the $L_R$ and $D_{KL}$ loss functions.

\begin{figure}[!htpb]
	\centering
	\begin{subfigure}[h]{0.23\textwidth}
		\includegraphics[width=\textwidth]{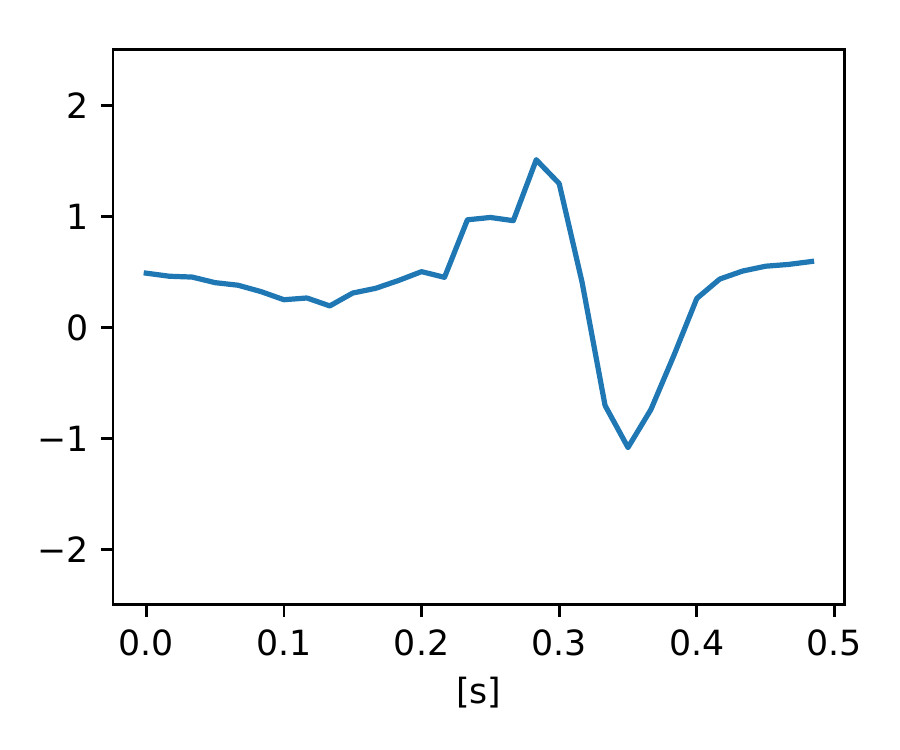}
		\caption{Embedding = [-2,2].}
		\label{fig:embmintwotwo}
	\end{subfigure}
	~
	\begin{subfigure}[h]{0.23\textwidth}
		\includegraphics[width=\textwidth]{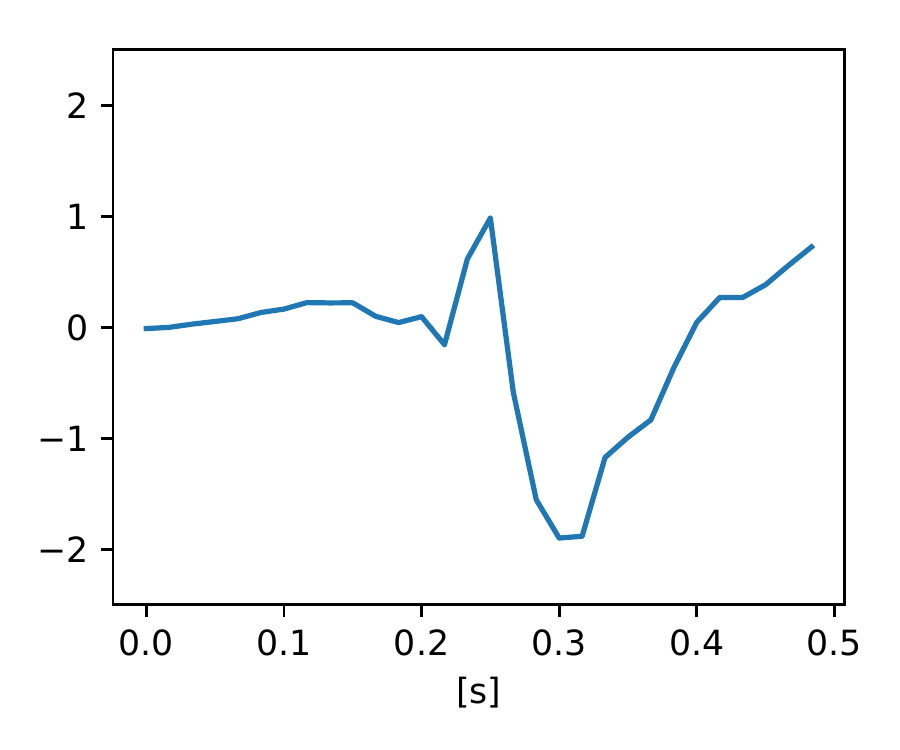}
		\caption{Embedding  = [2,2]}
		\label{fig:embtwotwo}
	\end{subfigure}
	~
	\begin{subfigure}[h]{0.23\textwidth}
		\includegraphics[width=\textwidth]{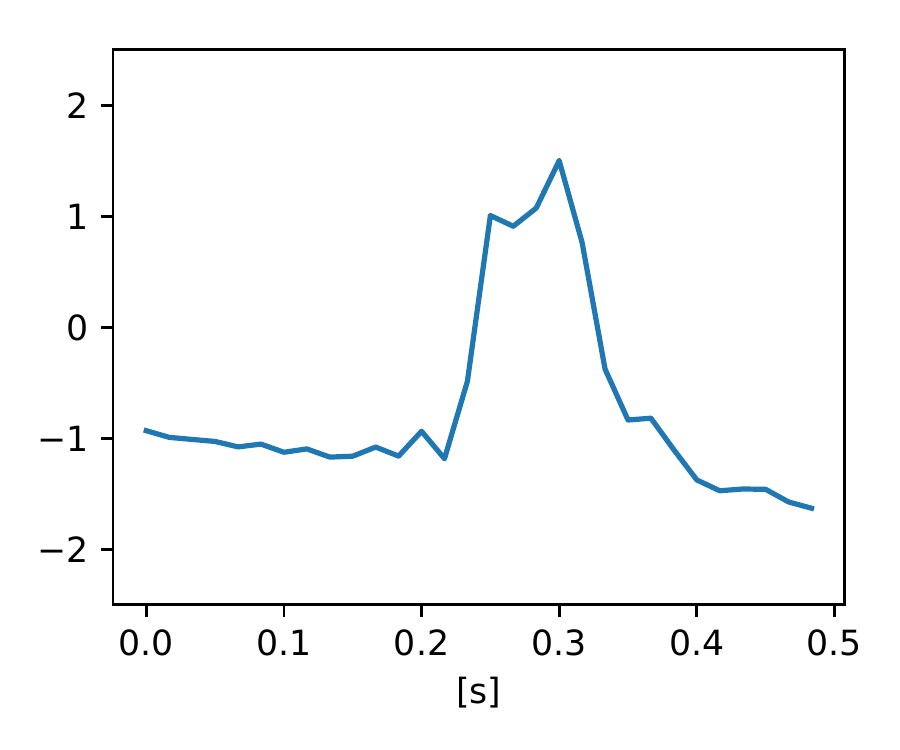}
		\caption{Embedding = [-2,-2]}
		\label{fig:embmintwomintwo}
	\end{subfigure}
	~
	\begin{subfigure}[h]{0.23\textwidth}
		\includegraphics[width=\textwidth]{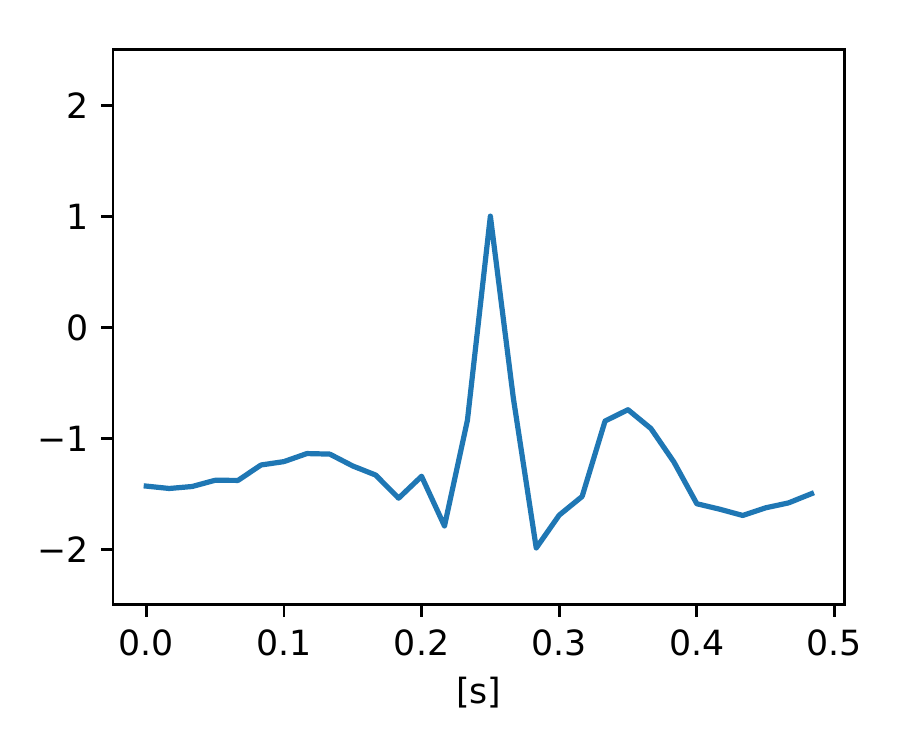}
		\caption{Embedding = [2,-2]}
		\label{fig:embtwomintwo}
	\end{subfigure}
	\caption{Decoding the edges of the embedding for the $\beta$-VAE model in the significant dimensions 0 and 9. The values of the other dimensions are fixed at zero.}
	\label{fig:embeval}
\end{figure}

With these two dimensions, the entire beat space can be visualized as shown in Fig.~\ref{fig:embeval} where the decoder is evaluated with embeddings at the four corners of the embedding space. Each beat can be represented as a smooth transition within this beat space and the position within this beat space indicates the prominence of specific beat features.

With the combination of a linear dense network and the $\beta$-VAE method, an independent, interpretable and explainable ECG beat embedding can be discovered. The embedding consists of several characteristic beats as base to setup the embedding space. This learned embedding space can subsequently be used in other models to enhance their interpretability and enable the use of machine learning models in clinical practice.



\section{Conclusion}
\label{sec:conclusion}
ECG beat classification is an important aspect of ECG analysis and is used in various branches of medicine. State-of-the-art neural network and deep learning models are capable of achieving a high classification accuracy. However, there is no human interpretable explanation for the classification decision of the model. By extending linear dense models to include a $\beta$-VAE embedding as illustrated in this work, representative beat patterns can be identified leading to an interpretable, explainable and independent embedding space. The resulting model is no longer black-box and beats can be represented as combinations of learned independent base beats. 

\section*{Acknowledgement}
This research received funding from the Flemish Government under the “Onderzoeksprogramma Artificiële Intelligentie (AI) Vlaanderen” programme.

\bibliographystyle{IEEEtran}
\bibliography{bibl}

\end{document}